\documentclass{article}
\usepackage{spconf,amsmath,graphicx}
\usepackage{algorithm}
\usepackage{graphicx}
\usepackage{amssymb}
\usepackage{algorithmic}
\usepackage{multirow}
\usepackage{colortbl}
\usepackage{booktabs}
\usepackage{newfloat}
\usepackage{listings}
\usepackage{amsmath}
\usepackage[colorlinks,linkcolor=blue]{hyperref}
\title{Parallel Augmentation and Dual Enhancement \\
for Occluded Person Re-identification}
\name{
    Zi Wang\textsuperscript{\rm 1},
    Huaibo Huang\textsuperscript{\rm 2}, 
    Aihua Zheng\textsuperscript{\rm 3}, 
    Chenglong Li\textsuperscript{\rm 3}, 
    Ran He \textsuperscript{\rm 2\thanks{Corresponding author: Aihua Zheng (ahzheng214@foxmail.com)}} 
}
\address{
    \textsuperscript{\rm1} School of Computer Science and Technology, Anhui University
    \textsuperscript{\rm2} MAIS \& CRIPAC, CASIA\\
    \textsuperscript{\rm3} IMIS Laboratory of Anhui Province, Anhui Provincial Key Laboratory of MCC, Anhui University
}
\begin{document}
%\ninept
%
\maketitle
\begin{abstract}
    Occluded person re-identification (Re-ID), the task of searching for the same person's images in occluded environments, has attracted lots of attention in the past decades. Recent approaches concentrate on improving performance on occluded data by data/feature augmentation or using extra models to predict occlusions. However, they ignore the imbalance problem in this task and can not fully utilize the information from the training data. To alleviate these two issues, we propose a simple yet effective method with \textbf{P}arallel \textbf{A}ugmentation and \textbf{D}ual \textbf{E}nhancement (\textbf{PADE}), which is robust on both occluded and non-occluded data and does not require any auxiliary clues. First, we design a parallel augmentation mechanism (\textbf{PA}M) to generate more suitable occluded data to mitigate the negative effects of unbalanced data. Second, we propose the global and local dual enhancement strategy (\textbf{DE}S) to promote the context information and details. Experimental results on three widely used occluded datasets and two non-occluded datasets validate the effectiveness of our method. The code is available at \href{https://github.com/littleprince1121/PADE_Parallel_Augmentation_and_Dual_Enhancement_for_Occluded_Person_ReID}{PADE (GitHub)}. 
\end{abstract}

\section{Introduction}
\begin{figure}[t]
    \centering
    \includegraphics[width=0.9\columnwidth]{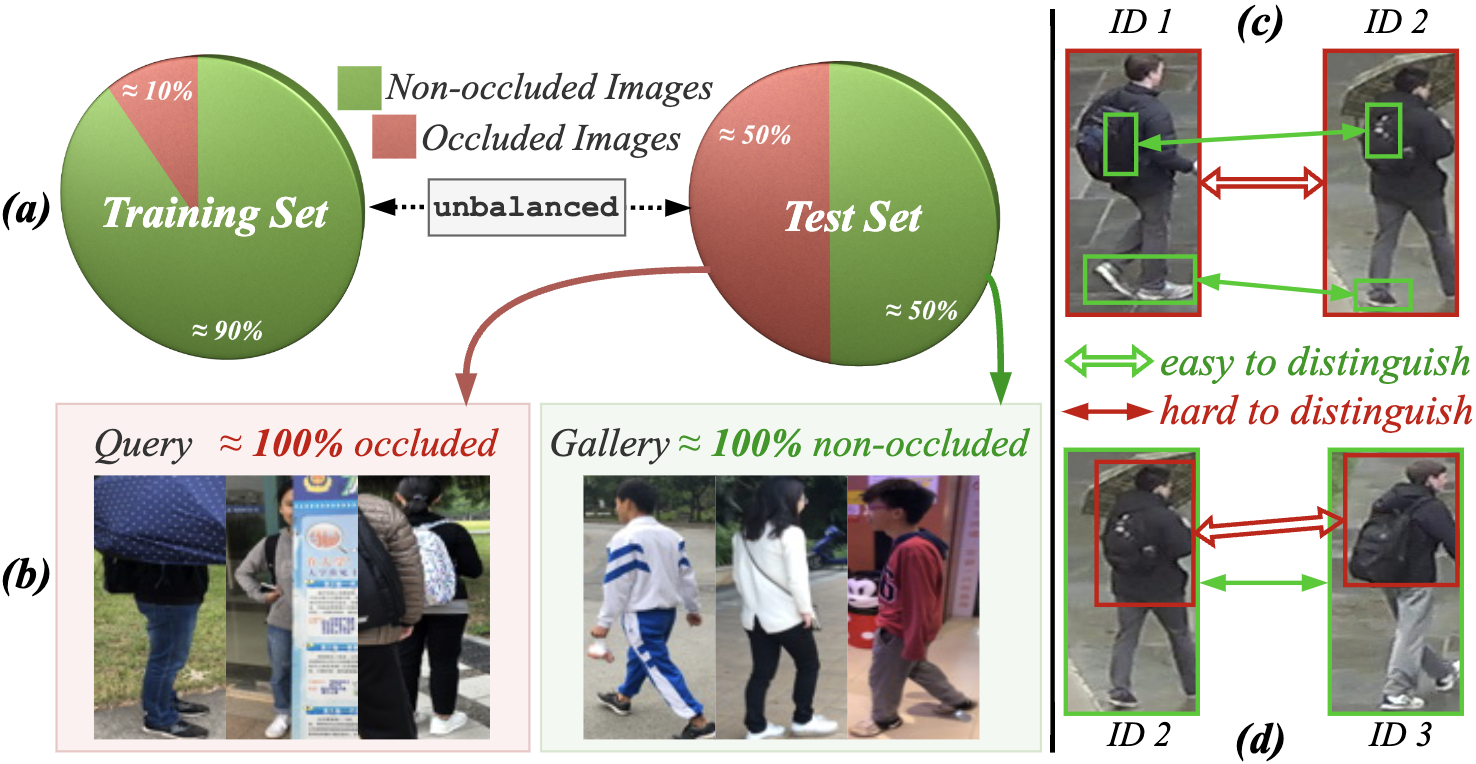} 
    \caption{(a) \& (b): Imbalance problem. (c) \& (d): Global and local information have their advantages, respectively.}
    \label{figure_motivation}
    \end{figure}

\begin{figure*}[t]
\centering
\includegraphics[width=1.9\columnwidth]{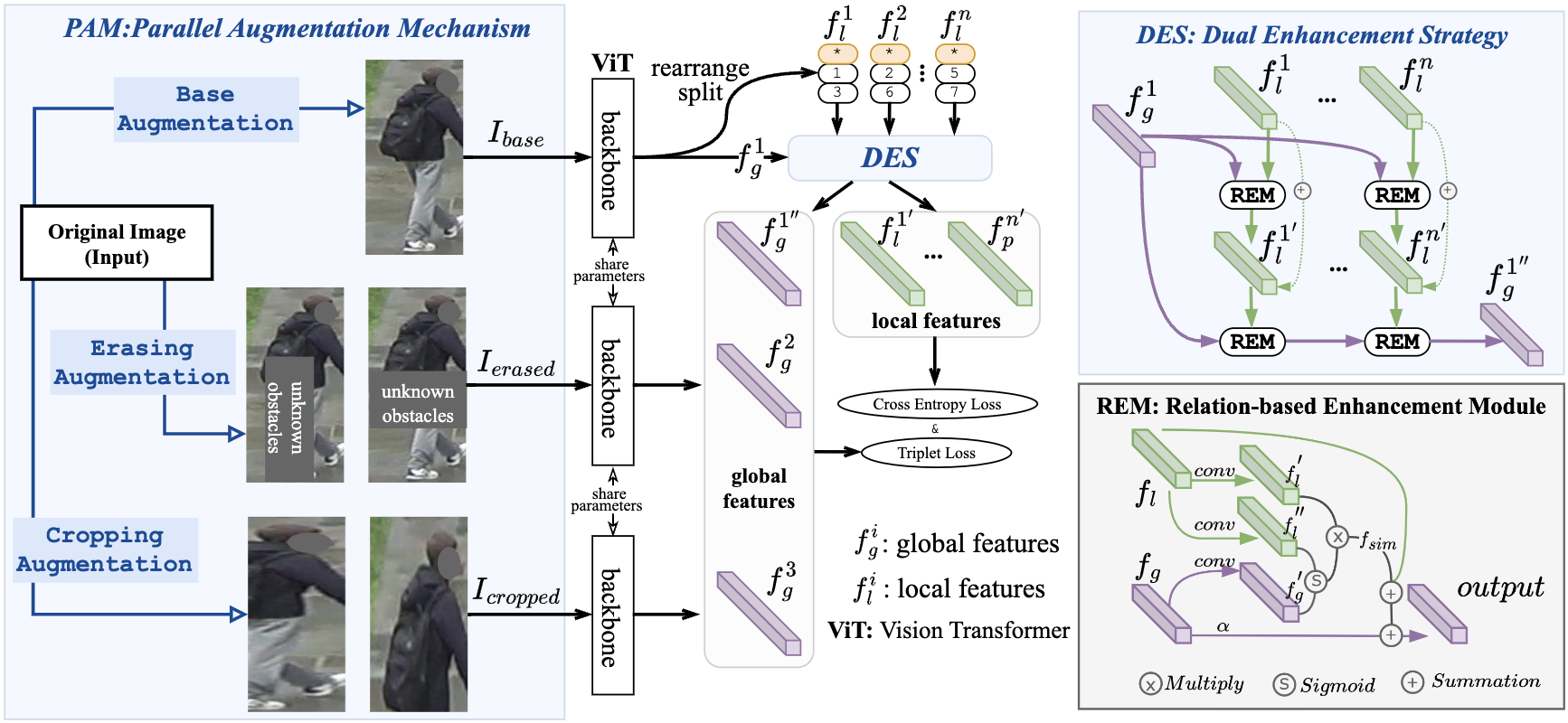} 
\caption{Overall structure of \textbf{PADE}. First, we implement erase and crop operations on original inputs to form the image triplet (Original, Erased, and Cropped images). The image triplet will be sent to the ViT-based backbone to extract global features. Then the global and local features from the non-occluded image branch will be interactively enhanced by each other. }
\label{figure_network}
\end{figure*}

Occluded person Re-ID, which incorporates the data obscured by various obstacles, has recently gained popularity.
And the occlusions are uncommon in the training set \cite{market1501, occDuke} while abundant in the test set (especially in query), as illustrated in Fig.~\ref{figure_motivation} (a). Training with such unbalanced data increases the challenge for the network while testing on unknown data. 
Efforts in data and feature augmentation are emerging to eliminate the imbalance between training and testing. Most methods \cite{transreid, PCB, OSNet, lei2008gaborFG} employ standard data augmentation such as random flipping, random deleting, random cropping, and so on. Furthermore, FED \cite{FED} provides feature augmentation strategies to improve the network's adaptability to occluded data. The widely used data/feature augmentation mechanisms take one image/feature as the input and output only one changed image/feature to the subsequent network for training. 
However, as illustrated in Fig.\ref{figure_motivation} (b), practically all occlusions occur in the query, and the gallery images almost have no obstructions in occluded Re-ID datasets \cite{partialREID, occREID}. The methods mentioned that focus on data/feature augmentations ignore the unbalanced occlusion between query and gallery. 
To increase the robustness of the network on both the non-occluded data (in the gallery) and the occluded data (in the query), we propose a data augmentation method called the Parallel Augmentation Mechanism (PAM). Our PAM consists of three independent components: Base Augmentation (BA), Erasing Augmentation (EA), and Cropping Augmentation (CA). In our parallel augmentation mechanism, EA only implements the erase operation, and CA only crops the original image. We will obtain an image triplet after the PAM, as shown in Fig. \ref{figure_network} (left). Then the ViT-based feature extractor takes the image triplet as the input.

% \begin{figure}[t]
% \centering
% \includegraphics[width=1\columnwidth]{augmentTypes.png} % Reduce the figure size so that it is slightly narrower than the column. Don't use precise values for figure width.This setup will avoid overfull boxes.
% \caption{(a) The traditional augmentation implements crop and erase on the images, and then outputs an augmented image. (b) For each input, we perform both the erasing and cropping augmentations in parallel and finally feed it into the feature extractor along with the non-occluded image.}
% \label{figure_augmentTypes}
% \end{figure}

Additionally, both details and context information are crucial for the Re-ID task. As illustrated in Fig. \ref{figure_motivation} (c), we can simply identify \textit{ID1} and \textit{ID2} by local details while finding it hard to distinguish them based on their outward appearance. \cite{MGCA, PVPM, PFD} propose using additional clues by leveraging foreground segmentation and pose estimation models. \cite{PCB, RGBNT201, IEEE} propose to split the global feature into several parts and use finer features with detailed information for training. In some cases, the global information becomes more crucial when the body is hindered by unknown impediments or the details are similar, as shown in Fig.\ref{figure_motivation} (d). ViT-based approaches \cite{transreid, FED, PFD} propose using ViT as a feature extractor due to the extreme sensitivity of global context information.
However, when encountering new datasets, the methods using auxiliary clues are susceptible and require extra annotations or fine-tuning. While the different persons have similar appearances, only using extracted global information is not discriminative enough. 
To make full use of global and local features, we propose the Dual Enhancement Strategy (DES) to enhance context information and details by forcing them to promote each other. As shown in Fig. \ref{figure_network}, the global and local features will be enhanced in two sequential steps. First, each local feature can be boosted by the context information in the global feature, and then the global feature can absorb the detailed information in the enhanced local features. It is worth noting that our DES does not require additional annotation or model assistance.
Our contributions are as follows: (1) We design the Parallel Augmentation Mechanism to form image triplets that will improve the robustness of the network. (2) We propose to enhance the global and local features in an interactive way according to the Dual Enhancement Strategy. (3) Our method does not need any auxiliary clues, and experimental results on five Re-ID datasets demonstrate the effectiveness and generality of the proposed methods.

\section{Proposed Methods}
    \subsection{Parallel Augmentation Mechanism}
    The widely used data augmentations are performed randomly and in a serial manner. However, occluded person Re-ID aims to match non-occluded and occluded images of the same person. To solve the unbalanced testing problem, we design the parallel augmentation mechanism (PAM), which generates augmented images in a parallel manner.
    For each original input image $I$, we implement three augmentations on it to obtain the image triplet [$I_{base}, I_{erased}, I_{cropped}$] for training. This process can be formulated as:
    \begin{equation}
        \begin{aligned}
            I_{base} = BA(I), I_{erased} = EA(I), I_{cropped} = CA(I),
        \end{aligned}
    \end{equation}
    where BA($\cdot$), EA($\cdot$), and CA($\cdot$) denote Base Augmentation, Erasing Augmentation, and Cropping Augmentation, respectively. Specifically, we keep the resize and normalization operations in all three augmentations. Compared to traditional data augmentation, the BA($\cdot$) only changes the size of the input image, the EA($\cdot$) only adds obstacles at random locations on the image, and the CA($\cdot$) only crops the image irregularly. After PAM, we obtain an image triplet, which consists of one image similar to the non-occluded image and two augmented images with different types of occlusion. Then, all three images will be sent to a parameter-shared multi-branch network:
    \begin{equation}
        \begin{aligned}
            f_g^1, f_g^2, f_g^3 = \theta(I_{base}, I_{erased}, I_{cropped}),
        \end{aligned}
    \end{equation}
    where $\theta(\cdot)$ denotes the feature extractor, we choose ViT-base \cite{ViT} as our backbone due to its powerful ability.

% \begin{listing}[tb]%
% \caption{Parallel Augmentation Mechanism}%
% \label{lst:PAM}%
% \begin{lstlisting}
% BaseAugmentation = Compose([
%     Resize(TrainSize),
%     ToTensor(),
%     Normalize()])
    
% EraseAugmentation = Compose([
%     Resize(TrainSize),
%     ToTensor(),
%     Normalize(),
%     RandomErasing(probability=1)])
    
% CropAugmentation = Compose([
%     Resize(TrainSize),
%     Pad(30),
%     ToTensor(),
%     Normalize(),
%     RandomResizedCrop(size=TrainSize)])
% \end{lstlisting}
% \end{listing}

    \subsection{Dual Enhancement Strategy}
        Many approaches extract critical global and local features in the training phase and optimize them together. However, they ignore the context information, and the details may not be valid at the same time. To solve this problem, we design the dual enhancement strategy (DES) to enhance both global and local features in an interactive way inspired by \cite{IEEE}.

        % \begin{figure}[t]
        %     \centering
        %     \includegraphics[width=0.9\columnwidth]{DES.png} % Reduce the figure size so that it is slightly narrower than the column. Don't use precise values for figure width.This setup will avoid overfull boxes.
        %     \caption{Illustration of dual enhancement strategy. }
        %     \label{figure_DES}
        % \end{figure}
        
        % \begin{figure}[t]
        %     \centering
        %     \includegraphics[width=0.9\columnwidth]{REM.png} % Reduce the figure size so that it is slightly narrower than the column. Don't use precise values for figure width.This setup will avoid overfull boxes.
        %     \caption{Structure of relation-based enhancement module. Take the single local feature $f_l$ to enhance the global feature $f_g$ for example.}
        %     \label{figure_REM}
        % \end{figure}
        
        After feature extraction, we can obtain the global features $f_g^1$ from $I_{base}$ and the local features ($f_l^1$, $f_l^2$ ... $f_l^n$) by splitting the $f_g^1$. The $n$ denotes the number of local features and is set to 4 in our experiment.
        First, each local feature will be enhanced by the global feature, and the original local features are treated as residuals and added back to the enhanced local features. This process can be formulated as follows:
        \begin{equation}
            \begin{aligned}
            f_l^{i'} = REM(f_l^i, f_g^1) + f_l^i, \quad i = 1, ..., n. \\
            \end{aligned}
        \end{equation}
        
        Then the global features are also enhanced similarly to absorb the details in the local features, as shown in Fig. \ref{figure_network}. This process can be formulated as:
        \begin{equation}
            f_g^{1''} =
                \begin{cases}
                  & REM(f_g^1, f_l^{i'}),\text{ if } i=1 \\
                  & REM(f_g^{1''}, f_l^{i'}),\text{ if } i=2,...,n.
                \end{cases}
        \end{equation}
        where the REM($\cdot$) denotes the relation-based enhancement module and can be formulated as:
        \begin{equation}
            \begin{aligned}
            & f_{sim} = \sigma(f_l^{'} \odot f_g^{'}) * f_l^{''}, 
            & f_g^{'} = f_{sim} + f_g + f_l,
            \end{aligned}
        \end{equation}
        where $\sigma$ denotes sigmoid, the $f_l^{'}$, $f_l^{''}$ and $f_g^{'}$ are the features after $Conv1 \times 1$, the $\odot$ denotes transpose multiply.
        The structure of REM is shown in Fig. \ref{figure_network}, and we use only one local feature and the global feature to illustrate. Then global features ($f_g^{1}$, $f_g^{2}$, $f_g^{3}$) and local features ($f_l^{1}$, ... $f_l^{n}$) will be concatenated to form the final person description.
        
        % The main difference of our dual enhancement strategy and interaction in \cite{IEEE} is that we consider both context and detailed information in person Re-ID, and the global and local features will promote each other in our DES. \cite{IEEE} only used the global information to enhance local features, ignoring the importance of local details.
        
        \subsection{Loss Function}
        We choose widely used cross-entropy loss ($L_{id}$) and triplet loss ($L_{tri}$) to train our model. All the global and local features are under the constraint of $L_{id}$ and $L_{tri}$. The final loss function can be formulated as:
        \begin{equation}
            \begin{aligned}
                L_{final} & = \sum_{i=1}^{3} L_{id}(p_g^{i}, y)
                          + \sum_{j=1}^{4} L_{id}(p_l^{j}, y)\\
                          & + \sum_{i=1}^{3} L_{tri}(f_g^{i})+  \sum_{j=1}^{4} L_{tri}(f_l^{j})
            \end{aligned}
        \end{equation}
        where the $p$ denotes the predicted results from global features $f_g$ and local feature $f_l$. The $y$ denotes the ground truth.

        \subsection{Implementation Details}
            The implementation platform of our experiment is Pytorch with RTX 3090Ti GPUs. The original learning rate is set as 0.008 and will be reduced in epochs 40 and 70. The max epoch is 170, the batch size is set to 32. We concatenate one global feature and four local features for testing, the dimension of the final person description is $768 * (1 + 4) = 3840$-dim. The Stochastic Gradient Descent (SGD) with a weight decay of 0.0004 is used in our experiment to fine-tune the whole network. Evaluation metrics are widely adopted Cumulated Matching Characteristics (CMC) curves and the mean Average Precision (mAP). 

        \begin{table}[t]
            \centering
        \resizebox{1\columnwidth}{!}{\begin{tabular}{|l|cc|cc|cc|} 
        \hline
        \multirow{2}{*}{\textbf{\textbf{\textit{Methods}}}} & \multicolumn{2}{c|}{\textbf{Occluded-Duke}} & \multicolumn{2}{c|}{\textbf{Partial-REID}} & \multicolumn{2}{c|}{\textbf{Occluded-ReID}}  \\ 
        \cline{2-7}                                                                                                         & mAP           & Rank-1                     & mAP        & Rank-1                       & mAP       & Rank-1                          \\ 
        \cline{1-7}
        HOReID* \cite{HOReID}                                                                                                         & 43.8          & 55.1                       & -          & 85.3                         & 70.2      & 80.3                            \\
        PVPM* \cite{PVPM}                                                                                                          & -             & -                          & 72.3       & 78.3                         & 61.2      & 70.4                            \\ 
        PFD* \cite{PFD}                                                                                                          & \textcolor{blue}{61.8}          & \textcolor{blue}{69.5}                       & -          & -                            & \textbf{\textcolor{red}{83.0}}      & {81.5}                            \\
        PCB \cite{PCB}                                                                                        & 33.7          & 42.6                       & 63.8       & 66.3                         & 28.9      & 41.3                            \\ 
        PGFA \cite{PGFA}                                                                                                     & 37.3          & 51.4                       & 61.5       & 69.0                         & -         & -                               \\
        OAMN \cite{OAMN}                                                                                                       & 46.1          & 62.6                       & 77.4       & 86.0                         & -         & -                               \\
        MoS \cite{MoS}                                                                                                     & 49.2          & 61.0                       & -          & -                            & -         & -                               \\
        ISP \cite{ISP}                                                                                                    & 52.3          & 62.8                       & -          & -                            & -         & -                               \\
        ViT Base \cite{ViT}                                                                                                   & 52.3          & 59.9                       & 74.0       & 73.3                         & 76.7      & 81.2                            \\
        PAT  \cite{PAT}                                                                                                    & 53.6          & 64.5                       & -          & \textcolor{blue}{88.0}                         & 72.1      & 81.6                            \\
        FED \cite{FED}                                                                                                          & 56.4          & 68.1                       & \textcolor{blue}{80.5}       & 83.1                         & 79.3      & \textcolor{red}{\textbf{86.3}}                            \\
        TransReID \cite{transreid}                                                                                                    & 59.2          & 66.4                       & -          & -                            & -         & -                               \\
        LoGoViT  \cite{phan2023logovit}                                                                                                   & 61.4          & 67.4                       & -          & -                            & -         & -                               \\
        \textbf{\textbf{PADE}} (ours)                                                                                   & \textbf{\textcolor{red}{63.0}} & \textbf{\textcolor{red}{72.3}}              & \textbf{\textcolor{red}{84.8}} & \textbf{\textcolor{red}{89.3}}                    & \textcolor{blue}{79.9} & {\textcolor{blue}{83.7}}                     \\
        \hline
        \end{tabular}}
            \caption{Experimental results on Occluded-Duke, Partial-REID, and Occluded-ReID (in $\%$). The best results are shown in \textbf{\textcolor{red}{blod}}, and the second results are shown in \textcolor{blue}{blue}. The superscript * denotes the method needs auxiliary clues.}
            \label{table_comparedSOTA}
        \end{table}

\section{Experiment}
    % \subsection{Datasets and Evaluation}
    % \label{sec_dataset}
    %     \textbf{Occluded-Duke} \cite{occDuke} contains 15618 training images, 2210 query images (occluded), and 17661 gallery images (some of them are occluded). \textbf{Partial-REID.} \cite{partialREID} contains 600 images of 60 IDs for testing, 300 images for query, and 300 images for the gallery. \textbf{Occluded-ReID} \cite{occREID}. The query set consists of 1000 occluded images, and the gallery set contains 1000 non-occluded images. \textbf{Market-1501} \cite{market1501} contains 1501 persons captured by 6 cameras. There are 12936 images (751 IDs) for training and 21960 images (750 IDs) for testing. \textbf{DukeMTMC-reID} \cite{dukeMTMC} contains 1404 persons captured in 8 camera views. It contains 16522 training images and 19889 testing images. \textbf{Evaluation metrics} are widely adopted Cumulated Matching Characteristics (CMC) curves and the mean Average Precision (mAP). 
    
    \subsection{Results on Occluded Re-ID}
    % We compared our method with state-of-the-art occluded Re-ID methods, including PCB \cite{PCB}, PGFA \cite{PGFA}, OAMN \cite{OAMN}, MoS \cite{MoS}, ISP \cite{ISP}, ViT Base \cite{ViT}, PAT \cite{PAT}, FED \cite{FED}, TransReID \cite{transreid}, and three methods using auxiliary clues HOReID \cite{HOReID}, PVPM \cite{PVPM}, PFD \cite{PFD}.

    We compared our method with state-of-the-art occluded Re-ID methods on Occluded-Duke\cite{occDuke}, Partial-REID\cite{partialREID} and Occluded-ReID\cite{occREID}.
    As shown in Table \ref{table_comparedSOTA}, our PADE achieves $ 63.0\%$/$84.8\%$ Rank-1 accuracy and $72.3\%$/$89.3\%$ mAP on Occluded-Duke and Partial-REID, respectively, and the results outperform all compared methods. Moreover, the results of our PADE on Occluded-ReID are better than all the methods with no auxiliary clues. Because of the additional keypoint detection model, PFD \cite{PFD} performs slightly better than ours on mAP evaluation, but the Rank-1 accuracy of PFD is still $2.2\%$ lower than ours. Our PADE considers the data unbalanced problems in testing and uses the parallel augmentation mechanism to improve the robustness of the network in the training phase. More visualization of datasets and ranking lists can be found at \href{https://github.com/littleprince1121/PADE_Parallel_Augmentation_and_Dual_Enhancement_for_Occluded_Person_ReID}{PADE (GitHub)}.

    \subsection{Results on non-occluded Re-ID}

    % We choose Market-1501 \cite{market1501} and DukeMTMC-reID \cite{dukeMTMC} to evaluate our PADE and several state-of-the-art methods, including PGFA \cite{PGFA}, OAMN \cite{OAMN}, MoS \cite{MoS}, ISP \cite{ISP}, ViT Base \cite{ViT}, JMLFNet\cite{zhang2023joint}, PAT \cite{PAT}, FED \cite{FED}, TransReID \cite{transreid}, HOReID \cite{HOReID}, PFD \cite{PFD} and LoGoViT \cite{phan2023logovit}.
    
    We choose Market-1501 \cite{market1501} and DukeMTMC-reID \cite{dukeMTMC} to evaluate our PADE and several state-of-the-art methods, and the results are shown in Table \ref{table_comparedNormal}. The dual enhancement strategy helps our PADE to enhance and utilize both context and detailed information and achieve high accuracy, we can observe PADE outperforms all the methods without using auxiliary clues and obtain $ 89.8\%$/$95.8\%$ mAP and Rank-1 on Market-1501. PADE even achieves the best results on Rank-1 accuracy compared with the methods using the extra pre-trained model on DukeMTMC-reID. 
    % The person images in non-occluded Re-ID datasets have fewer occlusions, so the context in global features and details in local features can provide more useful information. The dual enhancement strategy helps our PADE to enhance and utilize both context and detailed information and achieve high accuracy.

    \begin{table}[t]
        \centering
        \resizebox{1\columnwidth}{!}{
            \begin{tabular}{|l|c|cc|cc|} 
            \hline
            \multirow{2}{*}{\textbf{\textit{Methods}}} & \multirow{2}{*}{\begin{tabular}[c]{@{}c@{}}\textbf{\textit{Auxiliary}}\\\textbf{\textit{Clues}}\end{tabular}} & \multicolumn{2}{c|}{\textbf{Market-1501}} & \multicolumn{2}{c|}{\textbf{DukeMTMC-reID}}  \\ 
            \cline{3-6}
                                                       &                                                                           & mAP           & Rank-1                   & mAP           & Rank-1                      \\ 
            \hline
            HOReID \cite{HOReID}    & $\checkmark$                                              & 84.9          & 94.2                     & 75.6          & 86.9                        \\
            PFD\textsuperscript{$\dag$} \cite{PFD}         & $\checkmark$                                             & \textcolor{blue}{89.7}         & {95.5}                     & \textbf{\textcolor{red}{83.2}} & \textcolor{blue}{91.2}                        \\
            \hline

            PGFA \cite{PGFA}                                      & $\times$                                                 & 76.8          & 91.2                     & 65.5          & 82.6                        \\
            OAMN  \cite{OAMN}                                     & $\times$                                                 & 79.8          & 92.3                     & 72.6          & 86.3                        \\
            MoS \cite{MoS}                                       & $\times$                                                 & 86.8          & 94.7                     & 77.0          & 88.7                        \\
            ISP \cite{ISP}                                       & $\times$                                                 & 88.6          & 95.3                     & 80.0          & 89.6                        \\
            JMLFNet \cite{zhang2023joint}                                 & $\times$                                                & 89.2          & \textcolor{blue}{95.7}                     & 80.6          & 89.7                       \\
            PAT\textsuperscript{$\dag$} \cite{PAT}                                       & $\times$                                                 & 88.0          & 95.4                     & 78.2          & 88.8                        \\
            FED\textsuperscript{$\dag$} \cite{FED}                                        & $\times$                                               & 86.3          & 95.0                     & 78.0          & 89.4                        \\
            TransReID\textsuperscript{$\dag$} \cite{transreid}                                 & $\times$                                                & 88.9          & 95.2                     & 82.0          & 90.7                        \\

            \textbf{PADE\textsuperscript{$\dag$}} (ours)                              & $\times$                                                 & \textbf{\textcolor{red}{89.8}} & \textbf{\textcolor{red}{95.8}}            & \textcolor{blue}{82.8}          & \textbf{\textcolor{red}{91.3}}               \\ 
            \hline
            \end{tabular}}
        \caption{Experimental results on Market-1501 and DukeMTMC-reID (in $\%$). The superscript \textsuperscript{$\dag$} denotes the backbone of the method is ViT. The best results are shown in \textbf{\textcolor{red}{blod}}, and the second results are shown in \textcolor{blue}{blue}.}
        \label{table_comparedNormal}
    \end{table}

\begin{table}[t]
\centering

\scalebox{0.9}{\begin{tabular}{|c|c|c|c|c|c|} 
\hline
\multirow{3}{*}{\begin{tabular}[c]{@{}c@{}}{\textit{Ablation}}\\{\textit{Study}}\end{tabular}} & \multicolumn{3}{c|}{\textbf{\textit{Modules}}} & \multicolumn{2}{c|}{\textbf{Occluded-Duke}} \\ 
\cline{2-6}
 & \multicolumn{2}{c|}{\textbf{PAM}} & \multirow{2}{*}{\textbf{DES}} & \multirow{2}{*}{\textbf{mAP}} & \multirow{2}{*}{\textbf{Rank-1}} \\ 
\cline{2-3}
 & \textit{Cropping} & \textit{Erasing} &  &  &  \\ 
\hline
(a) & $\times$ & $\times$ & $\times$ & 57.3 & 67.8 \\ 
\hline
(b) & \checkmark & $\times$ & $\times$ & 60.1 & 69.5 \\ 
\hline
(c) & $\times$  & \checkmark  & $\times$ & 61.2 & 70.0 \\ 
\hline
(d) & \checkmark  & \checkmark  & $\times$ &  62.7 & 71.8 \\ 
\hline
(e) & \checkmark  & \checkmark  & \checkmark  & \textbf{63.0} & \textbf{72.3} \\
\hline

\end{tabular}}
\caption{Ablation study of the proposed method on Occluded-Duke (in $\%$). \textbf{PAM}: Parallel Augmentation Mechanism. \textbf{DES}: Dual Enhancement Strategy.}
\label{table_ablation}
\end{table}

    \subsection{Ablation Study}
    To prove the effectiveness of the proposed PAM and DES, we conduct the ablation study on Occluded-Duke by progressively introducing every component, as shown in Table \ref{table_ablation}. As shown in lines (b), (c), and (d) in Table \ref{table_ablation}, the model trained with PAM various occluded data achieves better results than the baseline (line (a)). In addition, DES utilizes the context and details information, further improving the accuracy of the proposed method, as Table \ref{table_ablation} (e) shows. Moreover, to evaluate the adaptability of the proposed PAM, we replaced the traditional augmentation in OSNet \cite{OSNet}, ViT base \cite{ViT}, and TransReID \cite{transreid} with PAM, the experimental results are shown in Table \ref{table_comparedAugment}. We can observe that all methods gain improvement after using the parallel augmentation mechanism (PAM). 

    \subsection{Robustness Evaluation on Occluded Data}
    To evaluate the robustness of PADE, we gradually increase the number of occlusion data by manually adding occlusions to the test data of DukeMTMC-reID \cite{dukeMTMC}. The crop operation and the erase operation are implemented on the original input to simulate occluded data. We can observe from Table \ref{table_occluded} that our PADE always achieves the best results and outperforms ViT base/TransReID on both mAP and Rank-1 accuracy. It means that our method suffers less when the occlusion data increases and our model is more robust on unexpected occlusions than the other two methods.
    
    \begin{table}[t]
        \centering
                \resizebox{0.85\columnwidth}{!}{
            \begin{tabular}{|l|l|l|} 
            \hline
            \multirow{2}{*}{\textbf{\textit{Methods}}} & \multicolumn{2}{|c|}{\textbf{Occluded-Duke}}   \\ 
            \cline{2-3}
                                                       & mAP                  & Rank-1                \\ 
            \hline
            OSNet \cite{OSNet} (base)                     & 29.5                 & 38.1                  \\
            OSNet \cite{OSNet} + \textbf{PAM}                     & 32.7 (\textbf{+3.2}) & 42.5 (\textbf{+4.4})               \\ 
            \hline
            ViT \cite{ViT} (base)                  & 52.3                 & 59.9                  \\
            ViT \cite{OSNet} + \textbf{PAM}                  & 57.9 (\textbf{+5.6}) & 66.2 (\textbf{+6.3})  \\ 
            \hline
            TransReID \cite{transreid} (base)                 & 59.2                 & 66.4                  \\
            TransReID \cite{transreid} + \textbf{PAM}                & 62.7 (\textbf{+3.5}) & 71.8 (\textbf{+5.4})  \\
            \hline
            \end{tabular}}
        \caption{Results of combining Parallel Augmentation Mechanism \textbf{(PAM)} with baseline on Occluded-Duke (in $\%$).}
        \label{table_comparedAugment}
    \end{table}

 \begin{table}[t]
\centering
\resizebox{1\columnwidth}{!}{\begin{tabular}{|cl|c|c|c|c|c|}
\hline
\multicolumn{2}{|c|}{Percentages}                                   & 20\%          & 40\%          & 60\%          & 80\%          & 100\%         \\ \hline
\multicolumn{1}{|c|}{\multirow{3}{*}{mAP}}    & ViT base \cite{ViT} & 57.2          & 53.6          & 50.9          & 48.0          & 45.5          \\ 
\multicolumn{1}{|c|}{}                        & TransReID \cite{transreid} & 60.6          & 58.4          & 56.2          & 54.2          & 50.2          \\  
\multicolumn{1}{|c|}{}                        & \textbf{PADE} (ours)      & \textbf{71.8} & \textbf{68.9} & \textbf{65.8}          & \textbf{63.5}          & \textbf{62.2}         \\ \hline
\multicolumn{1}{|c|}{\multirow{3}{*}{Rank-1}} & ViT base \cite{ViT} & 79.7          & 75.7          & 73.5          & 71.5          & 68.7          \\ 
\multicolumn{1}{|c|}{}                        & TransReID \cite{transreid} & 82.2          & 79.2          & 79.4          & 76.5          & 74.5          \\
\multicolumn{1}{|c|}{}                        & \textbf{PADE} (ours)     & \textbf{87.3} & \textbf{86.8} & \textbf{84.2} & \textbf{83.2} & \textbf{81.2} \\ \hline

\end{tabular}}
    \caption{Experimental results on different percentages of occluded data in DukeMTMC-reID (in $\%$).}
    \label{table_occluded}
\end{table}

\section{Conclusion}
    In this paper, we propose a simple yet effective method with Parallel Augmentation and Dual Enhancement for robust performance on both occluded and non-occluded person Re-ID. The parallel augmentation mechanism (PAM) can generate the image triplet (including non-occluded, erased, and cropped images), and help the network gain better ability on occluded-agnostic test data. The context information and details in global and local features will promote each other according to the dual enhancement strategy (DES). Both PAM and DES in our method can be flexibly embedded into other Re-ID methods, and they do not rely on any additional data annotations or models. In the future, we will focus on adaptive data augmentation and dynamic feature enhancement to deal with more complex occlusion environments. 

    \textbf{Acknowledgement.} This work was supported in part by the National Natural Science Foundation of China (Grants 62372003), the University Synergy Innovation Program of Anhui Province (Grant GXXT-2022-036), the Natural Science Foundation of Anhui Province (No. 2208085J18, No. 2308085Y40), the Natural Science Foundation of Anhui Higher Education Institution (No. 2022AH040014).
% ------------------------------------------------------------------------

\bibliographystyle{IEEEbib}
\bibliography{strings,refs}

\end{document}